# A Large-Scale Multi-Institutional Evaluation of Advanced Discrimination Algorithms for Buried Threat Detection in Ground Penetrating Radar

Jordan M. Malof[1], Daniël Reichman[1], Andrew Karem[2], Hichem Frigui[2], Dominic K. C. Ho[3], Joseph N. Wilson[4], Wen-Hsiung Lee[5], William Cummings[5], and Leslie M. Collins[1]

*Abstract*— In this paper we consider the development of algorithms for the automatic detection of buried threats using ground penetrating radar (GPR) measurements. GPR is one of the most studied and successful modalities for automatic buried threat detection (BTD), and a large variety of BTD algorithms have been proposed for it. Despite this, large-scale comparisons of GPR-based BTD algorithms are rare in the literature. In this work we report the results of a multi-institutional effort to develop advanced buried threat detection algorithms for a real-world GPR BTD system. The effort involved five institutions with substantial experience with the development of GPR-based BTD algorithms. In this paper we report the technical details of the advanced algorithms submitted by each institution, representing their latest technical advances, and many state-of-the-art GPR-based BTD algorithms. We also report the results of evaluating the algorithms from each institution on the large experimental dataset used for development. The experimental dataset comprised 120,000 $m^2$ of GPR data using surface area, from 13 different lanes across two US test sites. The data was collected using a vehicle-mounted GPR system, the variants of which have supplied data for numerous publications. Using these results, we identify the most successful and common processing strategies among the submitted algorithms, and make recommendations for GPR-based BTD algorithm design.

*Index Terms*—ground penetrating radar, landmine detection, buried threat detection

## I. Introduction

In this paper we consider the development of algorithms for the automatic detection of buried threats in ground penetrating radar (GPR) data. GPR is one of the most well studied and successful modalities for buried threat detection (BTD), and a large variety of BTD algorithms have been proposed in the literature for GPR-based BTD [1]–[18]. For example, GPR-based BTD algorithms have employed techniques from fields as varied as statistics [19]–[21], computer vision [22]–[24], and machine learning [6], [25]–[27].

Prolific development within the research community has advanced the effectiveness of GPR-based BTD systems, however most modern studies focus on proposing new algorithms, and they often compare their results against just one or two other algorithms [2], [22], [23], [28]–[31]. Systematic comparisons of modern algorithms are rare, and therefore it can be difficult to discern which algorithms, and more generally which processing practices, are best.

### A. A multi-institutional comparison of algorithms

In this work we report the results of a recent multi-institutional effort to develop, and compare, advanced buried threat detection algorithms. The effort involved five institutions with substantial GPR-based BTD experience: Duke University, University of Louisville, University of Missouri, University of Florida, and Chemring Sensors and Electronic Systems (CSES). A major objective of this effort was to identify the best processing approaches, and evaluate them in an unbiased manner, for potential inclusion in a real-world BTD system. As a result, each institution was provided with the same period of time for algorithm development; the same experimental dataset; and advanced knowledge of the scoring criteria.

The institutions were specifically tasked with developing *discriminators*. Discriminators must accept a small cube of GPR data (e.g., centered at a suspicious spatial "alarm" location), and return a decision statistic. The decision statistic indicates the relative likelihood that a buried threat is located at that location. An example cube of GPR data is shown in Fig. 1. The discriminators produced in this effort were compared using a large GPR dataset collected using a vehicle-mounted GPR-based BTD system, the variants of which have been involved in numerous publications over the preceding years [18], [23], [25], [32]–[36]. The GPR dataset collected using this system, and used to compare the discriminators, was comprised of 120,000 $m^2$ of surface area, collected over 13 lanes at two different US test sites, and encompassing 4,552 buried threat encounters.

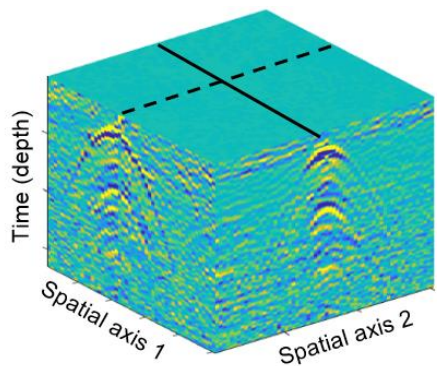

Fig. 1: Illustration of a cube of GPR data of the kind considered in this work. A slice of the data (i.e., an image) indicated by the solid line is projected onto Spatial axis 1, and the slice of data indicated by the dashed line is projected onto Spatial axis 2, for visualization of the contents of the GPR cube. Discrimination algorithms must distinguish between data cubes corresponding to threats and non-threats. This particular example corresponds to data collected at the location of a buried threat.

### B. Contributions of this work

The first major contribution of this work is a technical description of the discrimination algorithms submitted by each institution. While some algorithms represent the latest advances to previously proposed GPR-based BTD discriminators (e.g., gprHOG, EHD, LG), others are novel GPR-based BTD discriminators (e.g., SED). In either case, these algorithms incorporate the latest technical advances from each institution, and therefore represent state-of-the-art algorithms within the GPR-based BTD community.

The second major contribution of this work is a presentation and analysis of the results of the algorithm performance comparison. Notably, we compare and contrast the discrimination algorithms in order to distill the underlying processing strategies that are most effective for GPR-based BTD. Based upon these analyses we make recommendations for the design of effective BTD algorithms.

Our final contribution is comprised of two additional analyses of the results. The first analysis involved a rudimentary fusion of the algorithms, which yielded performance improvements. This result suggests that the discrimination algorithms provide complementary detection capabilities. Additional analysis also reveals relative advantages of some algorithms on shallow and deeply buried threats, respectively.

*C. Paper organization*

Section II describes the GPR system used to collect data for this study, and provides details regarding the experimental dataset. In Section III we discuss the experimental design that was used to evaluate the discriminators: the full detection processing pipeline for the GPR BTD system, data handling (e.g., how the data was used for training and testing supervised algorithms), and the scoring criteria. Section IV describes the technical details of the prescreening algorithm, which identifies suspicious locations for processing by the discrimination algorithms. Section V presents the technical details of the four discrimination algorithms submitted for the algorithm comparison. Section VI presents the experimental results. Section VII presents an analysis of the results, including fusion of the discriminator outputs, and a breakdown of their performance by target burial depth. Section VIII presents our conclusions, and recommendations for the design of effective GPR-based BTD algorithms.

II. THE EXPERIMENTAL SENSOR AND DATASET

In this section, we describe the vehicle-mounted radar system that was used to collect the GPR data used in our experiments, as well the GPR dataset that was collected using this system.

*A. The vehicle-mounted radar system*

The radar system employed to collect GPR data is comprised of an array of radar antennas that are mounted on the front of a vehicle. The array is oriented perpendicular to the direction of travel (i.e., cross-track), and the antennas are equally spaced along the array. During data collection, each antenna emits an ultra-wideband radar signal, consisting of a differentiated Gaussian pulse, and then measures the energy reflected back towards the array from the subsurface. The result of this procedure is a time-series of signal strengths, known as an A-scan [12], [23], from each antenna. As the vehicle moves forward down a path or lane, referred to as the "down-track" direction, each antenna records A-scans at regular down-track (spatial) intervals. The recording of cross-track A-scans (across the array), and down-track A-scans (along the direction of travel) results in a volume of GPR data. An example of a large GPR data volume is illustrated in Fig. 2.

*B. GPR dataset details*

The data in this experiment was collected at two distinct US testing sites, designated "Site A" and "Site B". A series of testing lanes (or pathways) were constructed at each site. A mixture of government manufactured and improvised threats were buried at known locations throughout each lane in order to evaluate the detection capabilities of the BTD system. The objects were buried at various depths, and contain varying levels of metal content. The burial depths and metal content of the target population is summarized in Table 1.

Site A was located in a temperate geographic location, and was comprised of 449 unique threats, distributed over 7 lanes. 56 total runs (or passes) were made over the lanes at Site A, resulting in a total of 48,000 $m^2$ of scanned lane area, and 3,368 threat encounters. Site B was located at an arid geographic location, and was comprised of 210 unique threats, distributed over 6 lanes. 34 total passes were made over the lanes, resulting in a total of roughly 72,000 $m^2$ of scanned lane area, and 1,184 threat encounters. In total, the testing data consisted of 90 runs over 13 unique lanes, encompassing roughly 120,000 $m^2$ of lane area.

Table 1: Metal content and burial depth of the experimental threat (encounter) population. The precise depth ranges for each burial depth category have been omitted to obscure the precise performance characteristics of the proprietary GPR-based BTD system. Here the "Deep burial depths" category corresponds to threats that are buried at (roughly) the $90^{th}$ percentile burial depth, or deeper.

|  | Metal | Low metal | Non-metal | Total |
|---|---|---|---|---|
| **Standard burial depths** | 1441 | 2121 | 465 | 4027 |
| **Deep burial depths** | 308 | 0 | 217 | 525 |

III. EXPERIMENTAL DESIGN

In this section we present details of the experimental design employed in this study to evaluate and compare the performance of the discrimination algorithms submitted by each institution. Importantly, this section also lays out the design specifications and/or constraints for the algorithms that were provided to each institution during development.

*A. The two-stage detection processing pipeline*

The full detection algorithm, or processing pipeline, employed on our BTD system is comprised of two sub-processes: *prescreening* and *discrimination* (i.e., classification). This processing pipeline is illustrated and described in Fig. 2, and has been applied in numerous previous studies that considered GPR-based BTD [1], [18], [23], [37]–[39]. The prescreener employed in our experiments was developed by the CSES Corporation, while the discrimination algorithms were developed by at least one of the other institutions (all of which are University research groups).

As discussed in Fig. 2, each discrimination algorithm must accept a cube of GPR data as its input, and return a real number indicating the relative likelihood that the location under consideration contains a buried threat. These cubes are centered at suspicious spatial locations, called alarms, that are identified by the prescreener. The cube of GPR data has a predetermined spatial extent, imposed by the need to process the data in real-time during system operation. Similarly, the radar system collects A-scans of a pre-determined, and fixed, (temporal) length. All algorithms were required to operate within these constraints.

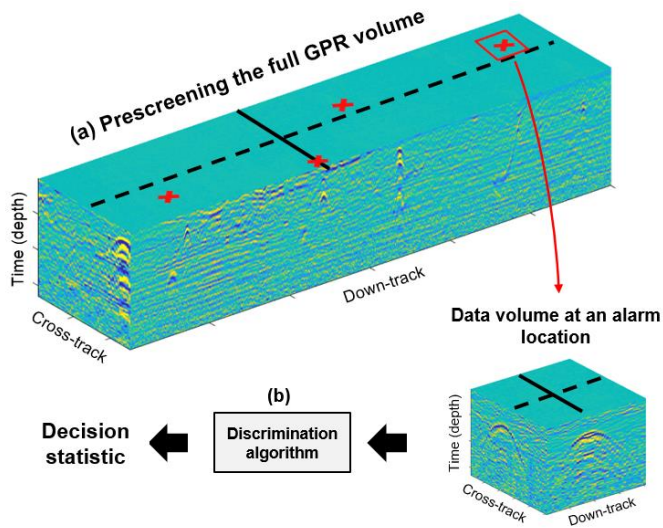

Fig. 2: Illustration of the two-stage detection processing approach employed in the BTD system in this study. (a) The first stage of processing is "prescreening", in which a relatively fast algorithm (a prescreener) is applied to the raw incoming volume of GPR data to identify a small set of suspicious locations, called alarms, for processing by the more computationally intensive discrimination algorithms. The prescreening operation reduces the amount of data considerably, making it possible to apply the discrimination algorithms in real-time (e.g., on a moving vehicle). (b) In the second stage of processing, discrimination algorithms are applied to cubes of GPR data extracted at each alarm location identified by the prescreener. The goal of the discrimination algorithm is to assign a "decision statistic" to each alarm location indicating the relative likelihood that the data corresponds to a true buried threat. Note that the discriminators operate only on the locations identified by the prescreener, and therefore any buried threats that are missed by the prescreener cannot be identified by the discriminators.

### B. Data handling: discriminator training and testing

All of the discriminators submitted for this comparison contain some type of supervised statistical, or machine learning, model. Supervised models have parameters that must be inferred using labeled examples of the classes of data we wish to categorize (e.g., threats and non-threats in this case). This process of parameter inference is often referred to as *training*, and it results in a trained algorithm that can then be applied to new, unlabeled, data in order to infer whether it is a threat or not (i.e., *threat inference*). Supervised learning models obtain excellent performance when sufficient training data is available, and they have now become popular in the GPR BTD literature [2], [18], [25], [38], [40], [41].

A common approach to assess the performance of any supervised algorithm (i.e., an algorithm containing a supervised model), is to employ cross-validation. This process involves training the models on some subset of the data, and then evaluating their ability to discriminate between the desired data classes on data that was excluded from training. In this work the discrimination algorithms were evaluated using a lane-based cross-validation approach, which we illustrate and describe in Fig. 3.

### C. Scoring the algorithms

The output of the discriminators (and prescreener) is a list of spatial alarm locations and their associated decision statistics. Before scoring these alarms, we must establish which of them should be considered correct alarms (i.e., they occur over real threats), and which of them are false alarms. In this work, any alarm located within a radius of 0.25m of a buried threat was considered to be a correct alarm, while all other alarms were considered false. This is a popular criterion that has been employed in numerous previous GPR-based BTD studies [30], [33], [42]–[45].

Given the true identity of each alarm (threat or non-threat), we can score the performance of the discriminators, and the prescreener. In this work we employ receiver operating characteristic (ROC) curves to measure performance. ROC curves quantify the tradeoff between the *correct* alarm rate (or probability of detection), $P_d$, and the *false* alarm rate (or probability of false detection), $FAR$, as we vary the sensitivity of the algorithm. The ROC curve is a popular metric in the GPR BTD literature [1], [6], [18], [23], [39], [46], where it is common to scale the $FAR$ metric so that it corresponds to the number of false alarms per square meter of scanned surface area. This representation of the BTD system's $FAR$ is often more interpretable and operationally relevant than false alarm probabilities. Unless otherwise stated, any reference to $FAR$ in this work refers to false alarms per square meter.

The vehicle-mounted BTD system considered in this work is proprietary, and therefore it was necessary to obscure its precise performance capabilities. In order to meet this need, while still effectively comparing the algorithms, we omitted all of the $FAR$ values from the ROC curves reported in this work (i.e., the values on the x-axis are omitted). Although $FAR$s were omitted, all of the ROC curves in this work (except Fig. 10 involving the prescreener) use exactly the same range of $FAR$ values on the x-axis. The presented range of $FAR$ values corresponds to those $FAR$s that were considered most operationally relevant for the discrimination algorithms. Consequently, the algorithm designers were tasked with developing discrimination algorithms to achieve the highest possible $P_d$ over this $FAR$ range.

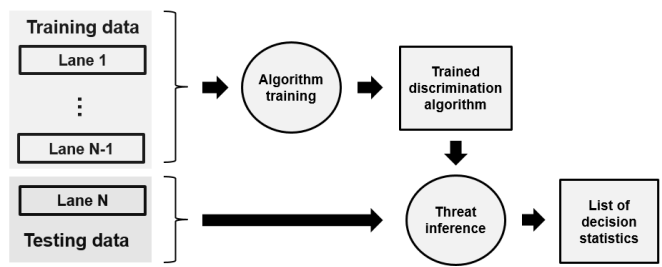

Fig. 3: Illustration of the lane-based cross-validation procedure used to evaluate the discrimination algorithms. Training of the supervised portions of the discrimination algorithms (i.e., parameter inference) is performed on prescreener alarms from $N-1$ lanes. This process yields a trained discrimination algorithm that can be deployed for threat inference on new, previously unseen, GPR data. The trained algorithm is applied for threat inference on prescreener alarms on the remaining lane that was withheld from training. This yields a list of decision statistics, indicating the relative likelihood that each of the prescreener alarms corresponds to a true buried threat. This process is performed $N$ times, so that each lane is employed exactly once for threat inference. The decision statistics from each lane are then aggregated and used to compute performance metrics for the algorithm.

## IV. THE PRESCREENING ALGORITHM

In this section we describe the prescreener algorithm, which is comprised of a fusion two individual prescreeners: the F2 prescreener, and the CCY prescreener. Each prescreener operates independently, and then their outputs are fused. First, we describe the two prescreeners, followed by the technique for fusing their outputs.

Note that, prior to prescreening, preprocessing is applied in which the time of the ground response is estimated at each spatial location, and then each A-scan is shifted so that the ground response occurs at the same time index across all spatial locations. Subsequently all data at, or above, the ground response is removed. These two preprocessing steps are common in GPR BTD [21], [29], [47], [48].

### A. The F2 prescreener

The F2 pre-screener is an updated version of F1 (Fast One) which in turn was derived from a LMS-based pre-screener developed by [42] to reduce its run time while maintaining comparable performance. The overall strategy of F2 is to identify locations in the GPR data with high signal energy, relative to the surrounding data. This strategy is implemented through a series of smoothing and constant false alarm rate (CFAR) processing [21], [49]. In this context, CFAR filtering usually refers to a process of statistically whitening data with a locally computed mean and standard deviation. With this in mind, F2 involves the following major processing steps:

(1) At each time index, median filtering the GPR volume in the down-track direction, to mitigate noise.
(2) At each time index, subtraction of the mean in the cross-track dimension, in order to remove panel-specific signal variations.
(3) Depth binning, whereby each A-scan is divided into non-overlapping bins (i.e., subsets of time indices). The set of A-scan values within each bin is replaced by the average of the top two values within that bin.
(4) CFAR filtering is applied along the time axis.
(5) The processed GPR volume is summed along the time axis, resulting in a single value at each spatial location (i.e., a 2-dimensional spatial map of intensities).
(6) 2-dimensional CFAR and Gaussian smoothing operations are applied, in that order, to the spatial map.

Alarms are obtained from the processed spatial map by applying thresholding to the resulting intensity values, and identifying connected components of pixels with intensities above the threshold.

### B. The CCY prescreener

To provide prescreening information complementary to that provided by F2 CSES developed a shape-based prescreener called Concavity (CCY). Given that the signal returns from real buried threats manifest as hyperbolas in the GPR data, the estimated convexity or concavity of signals in the data serve as a useful metric for identifying threats. This insight forms the basis for the CCY prescreener.

The CCY prescreener implementation begins by statistically whitening each point in the GPR volume based upon statistics computed on neighboring points at the same depth (time index). This step mitigates signal attenuation with respect to depth. Once this preprocessing is completed, a concavity measurement is computed at every spatial location in the GPR volume. This concavity measurement is itself computed from two concavity measurements: one computed on a down-track slice of GPR data, and one computed on a cross-track slice of GPR data.

Given either slice of GPR data, the concavity calculation proceeds in the same manner, and produces two concavity measures, $c^+$ and $c^-$ that are summed to obtain one concavity measure for the slice. A detailed description of the algorithm used to compute $c^+$ and $c^-$ is presented in Table 2, but we outline the algorithm here. The algorithm attempts to identify sequences of high magnitude pixels that form a concave curve. This search proceeds by considering the positive signals and the negative signals separately. For example, we consider only the positive signal by setting all of the negative pixel values to zero, and then identifying local maxima in the resulting image that exceed some value threshold. This processing is illustrated on real GPR data in Fig. 4 (middle image). $c^+$ is a measure of the concavity of the sequence of local maxima identified in the image. A similar process is applied to identify the negative signals to obtain $c^-$. The negative image maxima are also illustrated in Fig. 4.

The final estimate of concavity at a particular spatial location consists of summing the two concavity measures from the down-track slice (i.e., $c^+$ and $c^-$) with those from the cross-track slice (i.e., four total). Once a concavity measure is computed for every spatial location, a smoothing filter is applied to the resulting 2-dimensional map of concavity values. The CCY pre-screener reports as alarms the maximum points from 9x9 windows that are above a predetermined threshold.

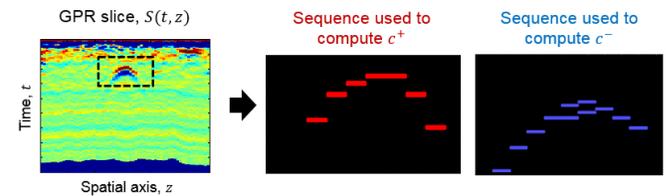

Fig. 4: (left) Illustration of a GPR slice. (middle) The sequences of local maxima values (in red) that were identified for computing $c^+$. This sequence was identified within the black dashed box in the GPR slice. The full procedure for identifying such sequences, and computing $c^+$, are described in Table 2. The sequences of local minima values (in blue) that were identified for computing $c^-$.

### C. Fusion of the prescreeners

Due to the complementary nature of the F2 (energy-focused) and CCY (shape-focused) prescreeners, it has been advantageous to prescreening performance to fuse their respective outputs. This can be accomplished by treating the alarms made by F2 and CCY as though they were generated from a single prescreener. However, it is likely that many pairs of alarms (i.e., one from F2 and one from CCY) will correspond to the same object (e.g., buried threat, subsurface rocks, roots, etc). This may result in wasteful processing of the same object via several different alarms, and create unnecessary vehicle stops or slowdowns if the object is perceived as a threat.

To address this problem, we merge alarms that are likely to be redundant. A simple and effective proxy for alarm redundancy is the relative proximity of the two alarms, and so alarms are merged using a proximity threshold. The exact details of the merging process are omitted here due to space considerations. In the event of a merger, however, the two alarms are usually replaced with a single alarm, and the decision statistic assigned to the new alarm is given by a weighted sum (i.e., an average) of the statistics from the individual alarms.

In general, the F2 and CCY decision statistics can differ by orders of magnitude, potentially interfering with the effectiveness of the weighted average, because the magnitude of one prescreener statistic dominates the other. To mitigate this problem, the CCY statistics are re-scaled as follows: a*CCY$^b$+c. In our experiments the three parameters, a, b, and c, were determined using training data to maximize the area under the ROC curve on the training data.

TABLE 2

| Algorithm to compute concavity measures, $c^+$ and $c^-$ |
|---|
| **Input**: A GPR slice, $S(t,z)$, where $t$ is the temporal axis and $z$ is the spatial dimension. Denote the spatial center of the patch by $z_0$. |
| **Parameters**: |
| $\omega$ = Size of the search window for maxima (positive integer) |
| $\gamma$ = The threshold to retain maxima values |
| **Output**: Two concavity measures, $c^+$ and $c^-$ |
| 1. Compute $S'(t,z) = |\max(S(t,z),0)| \ \forall (t,z)$ |
| 2. Compute $t^* = \max_i S'(i,z_0)$ |
| 3. If $S'(t^*,z_0) \geq \gamma$ set $c^+ = 0$ and go to step (9), **else,** initialize the set of coordinates X = $\{(t^*,z_0)\}$ |
| 4. **For** $j = 1 \ldots 5$ |
|   (a) Set $i^* = \underset{i \in [-\omega, \omega]}{\operatorname{argmin}} |t^* - i| \ s.t. \ S'(t^* + i, z_0 + j) \geq \gamma$ |
|   (b) **If** no $i^* = \emptyset$, go to step (5), |
|      **else** add $(t^* + i^*, z_0 + j)$ to X and set $t^* = t^* + i^*$ |
|   **End** |
| 5. Repeat step (4), with $j = -1 \ldots -5$. The set $C$ now contains a sequence of (spatially) neighboring points, with a maximum potential length of 11 points. See Fig. 4 for examples on real data. |
| 6. Construct X′ as the set of all possible sequences of consecutive points that can be constructed from the points in X. |
| 7. $c^+ = 1/|X'| \sum_{x \in C'} f(x)$, where $f(.)$ is a function that measures concavity as the difference between a sequence's mid-point and the average of its end-points. |
| 8. Set $S = -S$ and repeat steps (1)-(7) to obtain $c^-$ |

## V. DISCRIMINATION ALGORITHMS

This section presents the technical details of each of the discrimination algorithms submitted for the comparison. Although it was not a design constraint, all of the submitted discriminators follow the same basic processing pipeline, involving feature extraction and classification. This pipeline is illustrated and described in Fig. 5. For simplicity, we will refer to each discriminator by its feature extraction approach. For example, the first algorithm is based on the edge histogram descriptor (EHD) feature and we will therefore refer to it as "the EHD algorithm". Using the aforementioned nomenclature, a total of four discriminators were submitted for the comparison: EHD, Log-Gabor (LG), the Histogram of Oriented Gradients for GPR (gprHOG), and Spatial Edge Descriptors (SED). The description of each algorithm is broken down into three parts: feature extraction, classifier and training, and threat inference (i.e., how predictions are made on new data).

As described in Section III, all of these discriminators were required to operate on 3-Dimensional cubes of GPR data. All of the discriminators apply two initial pre-processing steps: (i) alignment of all A-scans so that the ground response in each A-scan occurs at the same time (depth) index; and (ii) the removal of all GPR data at, and above, the ground time index. LG only applies (i) for preprocessing.

All of the discriminators submitted to the comparison apply some form of depth-based calibration of the data. Although the precise approaches varied slightly, this calibration procedure always consisted of normalizing, or whitening, each pixel by removing a locally computed mean from the data, and dividing by a locally computed standard deviation. For example, at a given time index (depth), the mean and variance can be computed using a set of leading and trailing GPR samples, and subsequently used for whitening all pixels at that time index.

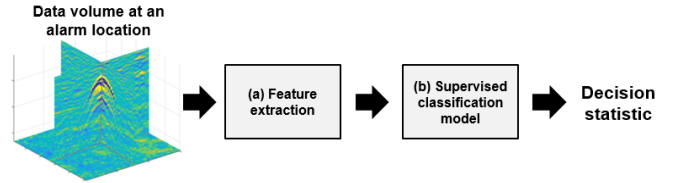

Fig. 5: Although not a design constraint, each discrimination algorithm submitted for the algorithm comparison adopted the same processing pipeline, illustrated here. (a) Feature extraction is the first step of processing, and consists of extracting of measures or statistics from the GPR data, with the aim of concisely (e.g., in a low dimensional vector) summarizing the important characteristics of the data. For example, many features attempt to encode the strength and orientation of edges in GPR slices. (b) The features extracted in (a) are provided to a supervised classification model that has been trained to distinguish between features corresponding to threats and non-threats, respectively. The classifiers assign a decision statistic to the input feature, indicating the relative likelihood that it corresponds to a threat.

### A. Edge Histogram Descriptors (EHD)

The EHD uses translation invariant features that are based on the histogram of edges in the GPR signature [32]. It is an adaptation of the MPEG-7 EHD feature [50] which encodes image texture information. It has been adapted to capture the spatial distribution of the edges within a 3D GPR volume. To keep the computation simple, 2D edge operators are used, and two types of edge histograms are computed. The first one, called EHD$^{DT}$, is obtained by fixing the cross-track dimension and extracting edges in the (time, down-track) plane. The second edge histogram, called EHD$^{CT}$, is obtained by fixing the down-track dimension and extracting edges in the (time, cross-track) plane.

*Feature extraction:* The EHD$^{DT}$ and EHD$^{CT}$ features are extracted from a GPR volume located at the prescreener alarm location with sizes (T,XT,DT) = (60, 15, 15). Let $S(t,x,y)$ denote this volume, and let $S_{ty}^{(x)}$ denote the x$^{th}$ plane of S. First, for each $S_{ty}^{(x)}$, four categories of edges are computed using 3x3 Sobel filters: vertical, horizontal, 45° diagonal, and 135° anti-diagonal. If the maximum of the edge strengths exceeds a preset threshold, the corresponding pixel is considered to be an edge pixel and is labeled according to the direction of the maximum edge. Otherwise, it is considered a non-edge pixel. Next, each $S_{ty}^{(x)}$ image is vertically subdivided into 7 overlapping sub-images $S_{ty,i}^{(x)}, i = 1, \ldots, 7$. For each $S_{ty,i}^{(x)}$, a 5 bin edge histogram, $H_{ty,i}^{(x)}$, is computed. The bins correspond to the 4-edge categories and the non-edge pixels. The EHD$^{DT}$ is defined as the concatenation of the 7 five-bin histograms. That is,

$EHD^{DT}(S(t,x,y)) = [\bar{H}_{ty,1} \quad ... \quad \bar{H}_{ty,7}]$, where $\bar{H}_{ty,i}$ is the cross-track average of the edge histograms of sub-images $S_{ty}^{(x)}$ over the 7 middle channels, i.e., $\bar{H}_{ty,i} = \frac{1}{7}\sum_{x=5}^{11} H_{ty,i}^{(x)}$. To compute the EHD output in the cross-track direction, i.e., $EHD^{CT}$, the y plane of S(t,x,y) is fixed and edges are extracted from $S_{tx}^{(y)}$ in a similar way.

*Classifier and training*: Support Vector Machine (SVM) [51] classifiers, with the radial basis function, were used for class prediction. One SVM for the $EHD^{DT}$ features ($SVM^{DT}$) is trained and a second SVM for the $EHD^{CT}$ ($SVM^{CT}$) is also trained. Both classifiers were implemented using the libSVM package [52]. All parameters were set to their default values. At each spatial location indicated by the prescreener, $EHD^{DT}$ and $EHD^{CT}$ features are extracted at multiple depths down the temporal axis by sampling one S(t,x,y) every 25 temporal indices. That is, we extract features from S(t=1..60, x, y), S(t=25..84,x, y), ... . This results in a total of 14 EHD features, $f_i$, i=1,…,14, per spatial location identified by the prescreener. For non-target alarms, any of the 14 features could be included in the training data. To maintain a balance between the number of training samples from both classes, we randomly select 5 of the 14 features. For targets, we use a kernel density estimator (KDE) to identify the few features that correspond to the most likely temporal location of the buried threat signal. Let $P^- = \{p_1^-, p_2^-, ... p_k^-\}$ be a set of k prototypes that summarize all of the non-target training alarms. Then, for each target training alarm, we estimate the KDE of its 14 EHD feature using

$$KDE(f_i) = \frac{1}{Z}\sum_{j=1}^{k} exp(-\beta\|f_i - p_j^-\|) \quad (1)$$

where β is a resolution parameter (learned during summarization of the non-target training alarms) and Z is a normalization factor. The $f_i$ features with very low KDE (close to zero) are selected as the most likely temporal locations that correspond to the actual target signatures and will be used for training.

*Threat inference:* For threat inference on new data, 14 $EHD^{DT}$ features and 14 $EHD^{CT}$ features are extracted from each prescreener alarm at multiple temporal locations. Then, the trained $SVM^{DT}$ and $SVM^{CT}$ classifiers are used to assign confidence values to each $EHD^{DT}$ and $EHD^{CT}$ feature respectively. Next, we fuse the decision statistics from both directions by taking their geometric mean at each temporal location. The final decision statistic is computed by summing the 3 top fused values.

## B. Log-Gabor (LG)

The Gabor filter, which is essentially a modulated Gaussian function at some frequency $f_o$, has been useful for many filtering tasks in signal processing. The Gabor filter bank is a series of Gabor filters created by imposing the constraint that the standard deviation governing the spread of the Gaussian function is inversely proportional to the modulation frequency $f_0$. Allowing the modulation frequency to increase in a dyadic manner creates the Gabor wavelets that are common for time-frequency signal analysis [53]. A distinct property of the Gabor filter is that its Fourier transform follows a Gaussian shape as well. Thereby the Gabor wavelets define a filter bank with each bandpass filter having a Gaussian shape frequency response.

In the processing of GPR data, the frequency spectrum of the radar echo reflected by a threat is asymmetric and has a long tail towards the high frequency region. To better capture the characteristics of the threat signal to aid detection, we apply the log-Gabor wavelets instead. The log-Gabor filter was first proposed by Field [54] in 1987 for image processing to better preserve the edge behavior in a natural image. The log-Gabor filter has a Gaussian frequency response in the log-frequency axis, thereby having a long tail response in frequency. Fig. 1 illustrates the difference in frequency responses between the Gabor and the log-Gabor filter.

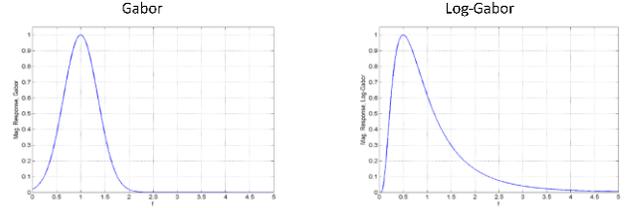

Fig. 6: Frequency response of the Gabor and the log-Gabor filter

*Feature extraction:* As noted previously, the threat signature often appears with a hyperbolic pattern in the B-scan. To capture the frequency as well as spatial responses, we shall apply the log-Gabor filters in 2-D for extracting the features for threat detection.[55] The 2-D log-Gabor filter is defined in the frequency domain through the polar coordinates. Let $\rho$ be the radius from the center and $\theta$ the angle from the x-axis. The frequency response of the log-Gabor filter response is given by

$$G_L(\rho,\theta) = exp\left(-\frac{1}{2[log(\sigma_\rho)]^2}\left[log\left(\frac{\rho}{\rho_0}\right)\right]^2\right) exp\left(-\frac{1}{2\sigma_\theta^2}(\theta - \theta_o)^2\right). \quad (2)$$

The parameters $\rho_o$ and $\theta_o$ control the location and the orientation of the filter response, and $\sigma_\rho$ and $\sigma_\theta$ determine the spreads in the frequency and angle. We create a set of 36 filters to represent the log-Gabor curvelets, by using 4 values of $\rho_o$ that decompose the frequency range and 9 values of $\theta_o$ that provide an angle resolution of 20 degrees. These 36 log-Gabor filters cover the frequency plane of a threat's GPR signal with 4 different filters at each of the 9 orientations. Fig. 7 shows the frequency as well as the spatial responses of the filters. The left part of the filters extract the rising edge behavior in the B-scan, the right part the trailing edge and the middle portion the horizontal.

The B-scan at the prescreener alarm location is separated into three spatial regions, with overlap: the left, middle and right. The left part of the image is processed by the filters in the first three columns in Fig. 7, the right part by the filters in columns 6 to 8 and the middle part by the filters in columns 4, 5 and 9. The filtering process is performed in the frequency domain. Each filter output is separated into 15 depth bins with 50% overlap. The element of the feature vector is the maximum of the energies over the 15 depth bins of each log-Gabor filter output.

In addition to applying the filters to the B-scan in the down-track, we also apply the log-Gabor filters to the 2-D image collected over the cross-track, as well as those from the positive-diagonal and anti-diagonal in the 3-D data cube at the alarm location. Each 2-D image results in 36 features. Since there are 4 total images, this yields a final feature vector of 144 total feature values.

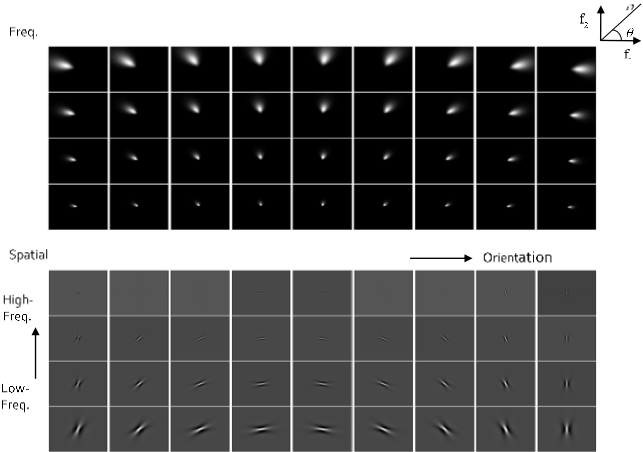

Fig. 7: Frequency and spatial responses of the 36 log-Gabor filters for feature extraction

*Classifier and training*: The feature vectors are used to train two SVM classifiers with an RBF kernel to perform target vs non-target classification. The first SVM is trained on the feature vector associated with each alarm. The second SVM is trained on a transformation of the feature vector. Before the max-operation is performed depth-wise, each alarm is represented by a matrix of feature vectors of dimension $15 \times 144$. The feature matrix is sorted in descending order of magnitude and the top 4 rows are used as individual feature vectors with the same label as that of the corresponding alarm. To avoid overtraining either SVM, a subset of the data is used for training. For the dataset of feature vectors for each SVM, the feature vectors of deeply buried threats are replicated once (because there are fewer deeply buried threats). Second, only 5% of the resulting samples are selected randomly to train the SVM classifier.

*Threat inference:* For threat inference, each SVM is applied to the feature vector transformation that corresponds to the data on which it was trained. Thus, the first SVM is tested on the feature vector that is the maximum value depth-wise for each of the 144 features for that alarm. The second SVM is tested on each of the 15 feature vectors for an alarm, resulting in 15 output confidences. The output confidences are sorted and the sum of the top 3 confidences is the final confidence for this classifier. The final confidence is the sum of the confidences of the two SVM classifiers.

### C. gprHOG: histogram of oriented gradients for GPR

Histogram of Oriented Gradients (HOG) is an image descriptor that was originally developed in the computer vision literature [56]. HOG was first applied for GPR-based BTD in [23] and has since become a very popular feature in the literature for BTD [2], [7], [28], [33], [57], [58]. Here we employ gprHOG, which incorporates several improvements to the original HOG implementation for GPR-based BTD in [23]. Further details and experimental justification for gprHOG can be found in the original paper [59].

*Feature extraction:* Before feature extraction, the data is downsampled in the temporal axis by a factor of 2, in accordance with [23] (although our numbers are slightly different due to differences in the radar system). The gprHOG feature is extracted on a cube of the pre-processed GPR data, denoted S(t,x,y), that is [T,XT,DT]=[18,18,18] in size. Similar to [23], we extract two gprHOG features: one cross-track feature, denoted $H_{tx}$, and one down-track feature, denoted $H_{ty}$. In [23], a single down-track B-scan was used to compute $H_{tx}$ and $H_{ty}$, respectively. With gprHOG however, $H_{tx}$ is an average of features computed over several neighboring cross-track B-scans, as illustrated in Fig. 8. This averaging step reduces noise in the feature and improves performance. Another important modification of HOG is the removal of the histogram (or block) normalization step. It was demonstrated in [33] that this substantially improves the effectiveness of the HOG descriptor for GPR-based BTD.

*Classifier and training*: In order to create a training dataset, we extract four image patches at threat locations using the MSEK algorithm [38]. MSEK identifies locations along GPR time axis that exhibit high levels of signal energy. At non-threat locations, 24 patches are extracted at regular intervals along the time axis. gprHOG features are extracted on all of the aforementioned patches and used for training. Another important improvement of the gprHOG algorithm (compared to [1]) is the use of two Random Forest classifiers [60]: one trained on downtrack gprHOG features, and one trained on cross-track gprHOG features. Both classifiers are trained with 100 trees.

*Threat inference:* At each prescreener alarm location, down-track and cross-track gprHOG features are extracted, respectively, at small regular intervals down the time axis. A final decision statistic is computed for each track by summing the top 12 classifier decision statistics. The two resulting statistics, one from cross-track and one from down-track, are multiplied to obtain a final statistic for the alarm.

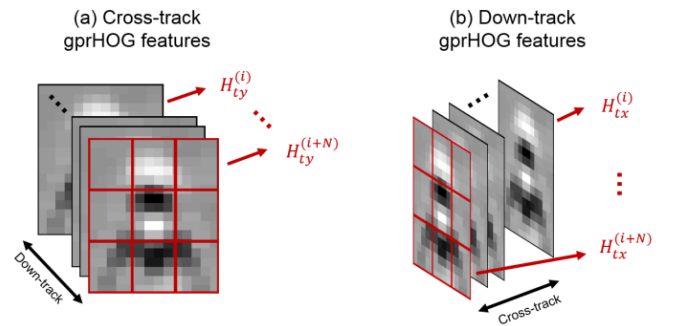

Fig. 8: Illustration of the extraction of the (a) cross-track gprHOG feature, denoted $H_{ty}$, and (b) the down-track gprHOG feature, denoted $H_{tx}$. $H_{ty}^{(i)}$ represents a HOG feature (with no block normalization) extracted on a time-crosstrack GPR slice located at the $i^{th}$ downtrack location within the GPR cube provided for feature extraction. The final cross-track gprHOG feature is given by the average of the the HOG features in each slice: $H_{ty} = (1/N) \sum_i H_{ty}^i$. Similarly, $H_{tx}^{(i)}$ represents a HOG feature extracted on a time-downtrack GPR slice located at the $i^{th}$ cross-track location within the GPR cube, and the final down-track feature is given by an average of the individual HOG features.

### D. Spatial edge descriptors (SED)

The SED algorithm is based upon extracting shape information, via gradient histograms, in 2-dimensional GPR images. Unlike most descriptors proposed for GPR-based

BTD, SED operates on spatial imagery: images comprised of GPR returns collected at the same instance in time. In these images, referred to as T-scans, buried threat signals appear circular rather than hyperbolic. This is illustrated in Fig. 9. The SED algorithm is designed to capture this shape to provide a descriptor of buried threats to the classifier.

*Feature extraction*: The SED feature is extracted from a GPR volume with $(T, XT, DT) = (50, 15, 15)$. The process for computing SED is illustrated in Fig. 9. For the $i^{th}$ temporal sample in the volume, $t_i$, we extract an image $I(x, y) = S(x, y, t = t_i)$. The image is divided into a $3 \times 3$ grid of cells. In each cell, a histogram of gradients is computed using four angle bins, resulting in a 36-dimensional descriptor for the image, $H_{xy}^{(i)}$. The final descriptor is constructed by averaging the descriptors over time, $H_{xy} = (1/T) \sum_i H_{xy}^i$. The averaging step is intended to increase the signal-to-noise ratio of the descriptor, due to uncertainty in the temporal location of the threat signal, and the tendency of the threat signal to appear over many time samples.

The gradients at each pixel are computed using four $3 \times 3$ pixel Sobel filters, each rotated by 45 degrees. An additional bin is added in this step corresponding to "no-edge" if the response to all templates is less than a specified threshold. In that case, a count is maintained of the number of pixels in the cell whose gradient response was less than the threshold. The threshold we use is 3, and this parameter has been relatively insensitive to dataset changes. We note that both using edge templates and a "no-edge" bin are similarly implemented in the EHD algorithm [32].

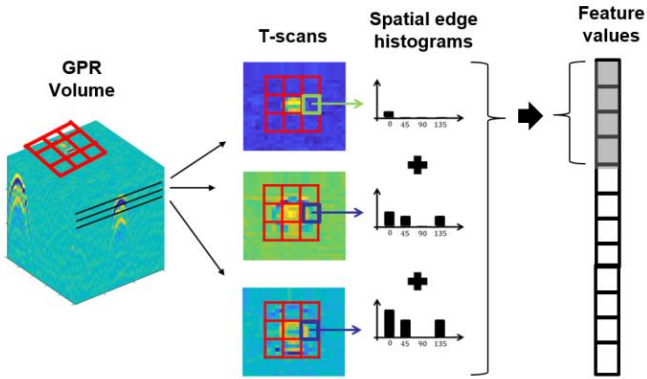

Fig. 9: Illustration of the process for computing the SED feature. Around the spatial coordinates of the prescreener alarm location and at the time-index estimated using MSEK, a $15 \times 15$ pixel patch is extracted which is divided into a $3 \times 3$ grid of cells. In each cell, a histogram of gradients is computed using 4 angle bins. The final descriptor vector supplied to the classifier is the average of computing SED on 50 temporal scans, with the first one computed at 5 temporal locations above the MSEK temporal location and the last one is 44 time samples lower.

*Classifier and training:* A Support Vector Machine (SVM) [51] with the radial basis function was used for class prediction. This was implemented using the libSVM package [52], with parameters $\gamma = 0.001$ and SVM cost parameter $C = 15$, which were chosen in cross-validation. At each spatial location indicated by the prescreener, the time index of the buried threat signal is estimated using an energy-based procedure referred to as MSEK [38]. The MSEK method computes the signal energy along an A-scan (i.e., over the GPR temporal dimension, smooths the energy time-series, and then identifies local maxima. For both threat and non-threat data, we extract *two* local maxima for training.

*Threat inference:* For threat inference on new data, SED features are extracted at regular intervals down the temporal (or depth) axis. The temporal locations are obtained, beginning with the first temporal index, and then by sampling one location every 25 temporal indices. This results in a total of 14 temporal locations. At each of these locations we extract SED features at each spatial location within a 5x5 spatial grid. This results in a total of 350 full SED feature vectors for each prescreener alarm. This extraction can be done very efficiently by reusing gradient computations between neighboring SED feature vectors. A final decision statistic for a given prescreener alarm location is computed by applying the classifier to all 350 SED feature vectors, and then summing the top 25 resulting decision statistics. We have found the performance to be largely insensitive to the number of decision statistics in the summation.

VI. EXPERIMENTAL RESULTS

A. *Prescreener results*

The performance results obtained by applying the prescreeners to the experimental GPR dataset are presented in Fig. 9. Each prescreener was applied to the same total area of lane, but obtained a different total number of false alarms, and so their respective ROC curves end at different *FAR* values. The *FAR* range of the x-axis in Fig. 9 has been extended so it includes the maximum *FAR* value across all of the prescreeners. Similarly, each prescreener missed a different total number of the true buried threats, and therefore each ROC curve reaches a different *maximum* $P_d$ value (which is obtained at its corresponding maximum *FAR* value).

The results in Fig. 9 indicate that CCY substantially outperforms F2 at all shared values of *FAR*. However, the fusion of the two prescreeners (referred to as the "Fusion" prescreener in Fig. 9) obtains a much greater $P_d$ value than either F2 or CCY at all shared values of *FAR*. The relative advantage of CCY over F2 suggests that it is important to leverage shape information in threat detection algorithms. CCY relies primarily (but not exclusively) on shape-based cues in the GPR data, while F2 relies primarily (though not exclusively) on signal energy. The large performance improvement yielded by their fusion demonstrates that, while a shape-focused approach has an overall advantage in our experiments, both energy and shape content appear to be important to obtain the best performance. This is implied by the substantial performance gain when fusing the two prescreeners.

Another important finding in Fig. 9 is that the Fusion prescreener obtains a greater *maximum* $P_d$ than either F2 or CCY. This implies that the F2 prescreener identified some threats that were not identified by CCY, further corroborating the complementarity of CCY and F2. The discrimination algorithms, even if they perform perfectly ($P_d = 1$ with no false alarms) can never identify buried threats that were not already identified by the prescreener. As a result, the maximum $P_d$ of a prescreener can be an important performance criterion, and one which is improved via the fusion of CCY and F2.

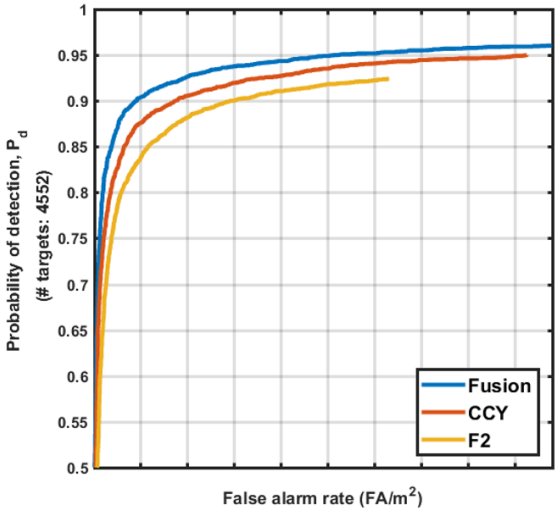

Fig. 10: ROC curves for the two individual prescreeners, F2 and CCY, as well as their fusion. Note that the y-axis has been truncated to the range $P_d = [0.5,1]$.

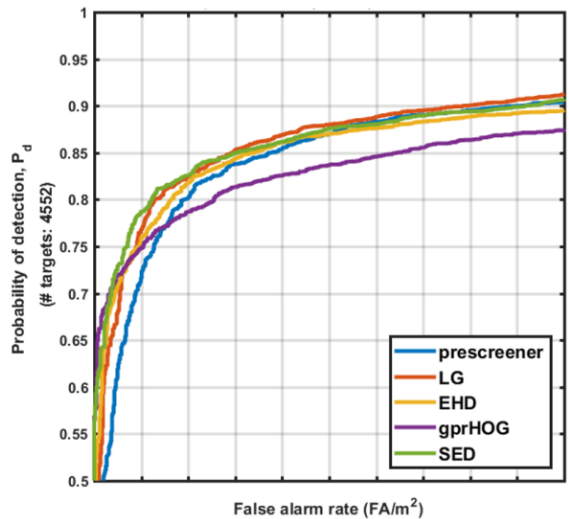

Fig. 11: ROC curves of all discrimination algorithms computed using in lane-based cross-validation on the (fusion) prescreener alarms. Note that the y-axis has been truncated to the range $P_d = [0.5,1]$.

*B. Individual discriminator results*

This section presents the results of evaluating the discrimination algorithms using lane-based cross-validation (Section III.B) on the alarms extracted by the fusion prescreener. The decision statistics of the fusion prescreener alarms were thresholded at a pre-determined operating sensitivity threshold, yielding 4,372 threat locations (96.1% of the total present). The total number of non-threat locations (i.e., false alarms) is omitted to obscure the precise performance capabilities of the system. The cross-validation results for all discrimination algorithms that were applied to this dataset are presented in Fig. 11.

The results indicate that no single algorithm consistently performs best (i.e., provides the best detection rate, $P_d$) across all of the *FAR* values. In the lower *FAR* range SED performs the best, while in the higher *FAR* range, LG performs best. EHD never provides the best $P_d$ but it provides comparable performance to SED and LG over most of the considered *FAR* range. The gprHOG algorithm briefly performs best at very low *FAR*, but then performs poorly for the remaining *FAR* range, providing lower $P_d$ than even the prescreener at most *FAR*s.

It is not surprising that the algorithms obtain (relatively) similar performance, due to the similarity in their processing strategies. The algorithm designs have gradually converged over time through the adoption of practices that have proven to be generally effective. For example, all of the algorithms use the two-stage processing pipeline outlined in Fig. 5, comprised of feature extraction and classification. Regarding feature extraction, each algorithm extracts features on GPR imagery (e.g., slices of the GPR volume) rather than the 3D cubes directly. In addition, all of the features are designed to encode shape information in the imagery, especially the hyperbolic pattern that is commonly associated with buried threats. Finally, all of the features involve aggregating the extracted shape information over large spatio-temporal regions of the GPR data (aside from gprHOG, which employs relatively smaller regions). Within classification, with the exception of gprHOG, all of the approaches also use the same classifier: an SVM with a radial basis function kernel.

## VII. FURTHER ANALYSIS

*A. Fusion experiment*

In this section we attempt to assess the relative complementarity of the discrimination algorithms by measuring the performance of a simple fusion of their outputs. Let $t_i$ be the decision statistic of the $i^{th}$ discriminator; we use a fusion comprised of a simple unweighted multiplication of *all* four of the discriminator decision statistics: $t_{fusion} = \prod_{i=1}^{4} t_i$. Before the fusion, we apply the popular Platt scaling [61] to the decision statistics of each discriminator. The Platt scaling applies a logistic regression (two parameters) to the statistics of each discriminator, after which $t_i \in [0,1]$, and the statistics approximate a class posterior probability. Unique scaling parameters were inferred for each algorithm using the decisions statistics generated by cross-validation. In order to minimize positive bias in the results, only the statistics from a single lane were used for parameter inference (i.e., training), and we note that the results were insensitive of the lane chosen for training.

After this scaling, $t_{fusion}$ can be interpreted as an "AND" operation between the discriminators, in which an object is only labeled as a threat if all of the individual discriminators label it as a threat. The results of this fusion are presented in Fig. 12. The results indicate that this simple fusion yields a substantial improvement in $P_d$ across all values of *FAR*. The substantial benefit of fusion, without any subselection or weighting of the individual discriminators, suggests that the algorithms possess some complementary decision characteristics, despite utilizing seemingly similar processing approaches.

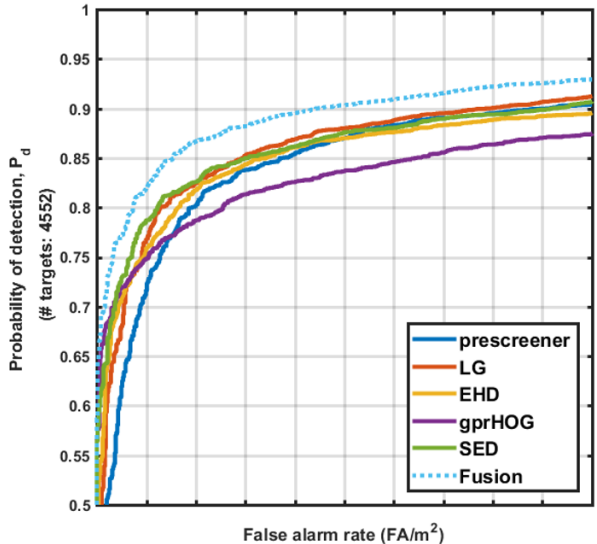

Fig. 12: ROC curves of all the individual discrimination algorithms, and their fusion via a simple unweighted geometric mean. Note that the y-axis has been truncated to the range $P_d = [0.5, 1]$.

### B. Algorithm performance by burial depth

In this section we evaluate the performance of the individual discrimination algorithms when measuring their performance on two disjoint populations of buried threats: threats that are buried at standard burial depths, and those with relatively deep burial depths. The definition of these categories is described in Table 1. Note that the supervised portions of the algorithms were not re-trained in each case; instead we computed ROC curves on the subsets of buried threats with the designated burial depths. These results are presented in Fig. 13 (standard burial depths) and Fig. 14 (deep burial depths).

As would be anticipated, the results indicate that the deeply buried threats are substantially more difficult to detect than those with a more shallow burial depth. This result is consistent with findings in the literature [1], [7], and is likely caused by the lower expected signal-to-noise ratio among deeply buried threats. It may also be exacerbated by the poor representation of deeply buried threats in the training dataset.

Interestingly however, the LG algorithm offers a substantial performance advantage over the other discrimination algorithms on deeply buried threats. This implies that, in addition to lower signal-to-noise ratios, deeply buried threats may also exhibit different signal characteristics that are captured more effectively by the LG features. One unique characteristic of the LG feature is that it encodes shape content at multiple scales, which may make it well suited to the smaller and/or weaker signals typical of deeply buried threats. Further investigation is needed to confirm this hypothesis however.

Among threats buried at common depths, the SED algorithm performs best over most of the ROC curve, and gains a noticeable performance advantage over LG. EHD also gains performance relative to LG in this regard. This is consistent with the hypothesis that threats may require different processing depending upon their burial depth. This processing may not only involve unique features for each burial depth, but it may also require resampling the training data. For example, it may be beneficial to replicate deeply buried threats when training any supervised classifier for application to that population.

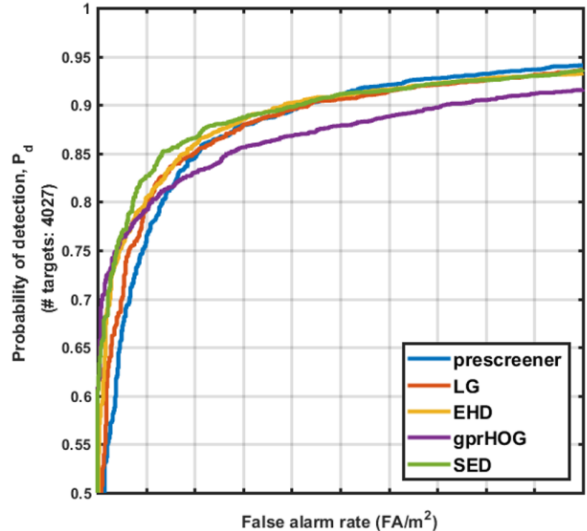

Fig. 13: ROC curves of all the individual discrimination algorithms when applied (but not trained) to threats buried at standard burial depths (i.e., not deep). Note that the y-axis has been truncated to the range $P_d = [0.5, 1]$.

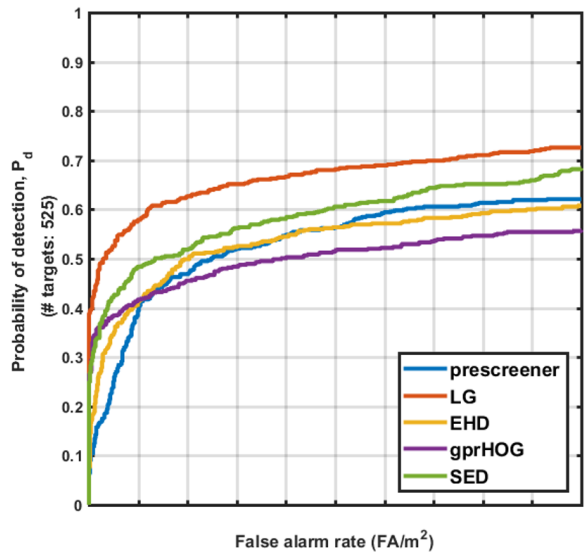

Fig. 14: ROC curves of all the individual discrimination algorithms when applied (but not trained) on threats buried at deep burial depths. Note that the y-axis *has not been truncated* to the range $P_d = [0.5, 1]$.

## VIII. CONCLUSIONS

In this work we report the results of a recent multi-institutional effort to develop, and compare, buried threat detection algorithms. The effort involved five institutions with an established track record in GPR-based BTD: Duke University, University of Louisville, University of Missouri, University of Florida, and Chemring Sensors and Electronic System (CSES). A major objective of this effort was to identify the best processing approaches, and evaluate them in an unbiased manner, for potential inclusion in a real-world BTD system. As a result, each institution was provided with the same period of time for algorithm development; the same experimental dataset; and advanced knowledge of the experimental design, and performance metrics.

The institutions (excluding CSES) were specifically tasked with developing *discriminators*. Discriminators must accept a small cube of GPR data (e.g., centered at a suspicious spatial location), and return a decision statistic, indicating the relative likelihood that a buried threat is located at that location. The discriminators produced in this effort were compared using a large GPR dataset collected using a vehicle-mounted GPR BTD system. The GPR dataset collected using this system, and used to compare the discriminators, was comprised of 120,000 m$^2$ of surface area, collected over 13 lanes at two different US test sites, and encompassing 4,552 buried threat encounters.

### A. Conclusions of the experimental results

The results reveal similar performance among most of the algorithms, with the SED and LG algorithm providing the best performance over large subsets of the considered range of false alarm rates (*FAR*). Further analysis revealed that the LG algorithm had a substantial advantage over the other algorithms on deeply buried threats, while SED most often outperformed other algorithms over buried threats with common burial depths (i.e., not deeply buried threats). A simple fusion of the algorithms, involving no sub-selection or weighting of the individual algorithms, yielded substantial performance improvements. This result suggests that the algorithms provide complementary detection capabilities, despite having many similar design characteristics.

### B. Conclusions regarding algorithm design

In this work we also provided a technical overview of the discriminators that were submitted by each institution. The algorithms employed many common designs, which provide insight into good design practices for BTD algorithms:

- Each algorithm employed two-stage processing comprised of feature extraction and classification.
- Each algorithm treats the GPR data like imagery, extracting features from slices of the GPR data rather than from 3D cubes directly.
- All of the feature extractors encode some type of shape information in GPR imagery, and aggregate this information over large spatio-temporal regions.
- Most of the approaches used an SVM classifier with a radial basis function kernel.
- It may be beneficial to apply different processing strategies to deeply buried threats and those at common burial depths (see discussion in Section VII.B).

In addition to these common practices, there were also many differences between the algorithms. Section VII provided experimental evidence that there is complementarity in the algorithms, and further effective practices may be revealed in the future by distilling which unique designs of each algorithm give rise to their respective advantages.

ACKNOWLEDGEMENTS

The authors would like to thank M. Scholl and P. Howard for their support of this work. This work was supported by the U.S. Army RDECOM CERDEC Night Vision and Electronic Sensors Directorate, via a Grant Administered by the Army Research Office under Grant W911NF-06-1-0357, Grant W911NF-13-1-0065, and Grant W911NF-14-1-0589.